\documentclass[10pt,twocolumn,letterpaper]{article}

\usepackage{iccv}
\usepackage{times}
\usepackage{epsfig}
\usepackage{graphicx}

\usepackage{soul}
\usepackage{url}
\usepackage[utf8]{inputenc}
\usepackage[small]{caption}
\usepackage{amsmath}
\usepackage{amsthm}
\usepackage{booktabs}
\usepackage{algorithm}
\usepackage{algorithmic}
\usepackage[switch]{lineno}

\usepackage{amsthm,amsmath,amssymb}
\usepackage{subfigure}
\usepackage{tabularx}
\usepackage{multicol}
\usepackage{multirow}
\usepackage{authblk}

\usepackage[pagebackref=true,breaklinks=true,letterpaper=true,colorlinks,bookmarks=false]{hyperref}
\iccvfinalcopy 

\def\iccvPaperID{11766} 

\ificcvfinal\fi

\begin{document}

\title{Bold but Cautious: Unlocking the Potential of Personalized Federated Learning through Cautiously Aggressive Collaboration}

\author[1]{Xinghao Wu}
\author[1,2]{Xuefeng Liu}
\author[1,2,3]{Jianwei Niu \thanks{Corresponding author}}
\author[1]{Guogang Zhu}
\author[4]{Shaojie Tang}
\affil[1]{State Key Laboratory of Virtual Reality Technology and Systems, School of Computer Science and Engineering, Beihang University, Beijing, China \authorcr \begin{small} \{wuxinghao, liu\_xuefeng, niujianwei, buaa\_zgg\}@buaa.edu.cn \end{small}}
\affil[2]{Zhongguancun Laboratory, Beijing, China}
\affil[3]{Zhengzhou University Research Institute of Industrial Technology, School of Information Engineering, Zhengzhou University, Zhengzhou, China}
\affil[4]{Jindal School of Management, University of Texas at Dallas, USA \authorcr \begin{small} shaojie.tang@utdallas.edu \end{small}}

\maketitle
\ificcvfinal\thispagestyle{empty}\fi


\begin{abstract}
Personalized federated learning (PFL) reduces the impact of non-independent and identically distributed (non-IID) data among clients by allowing each client to train a personalized model when collaborating with others. A key question in PFL is to decide which parameters of a client should be localized or shared with others. In current mainstream approaches, all layers that are sensitive to non-IID data (such as classifier layers) are generally personalized. The reasoning behind this approach is understandable, as localizing parameters that are easily influenced by non-IID data can prevent the potential negative effect of collaboration. However, we believe that this approach is too conservative for collaboration. For example, for a certain client, even if its parameters are easily influenced by non-IID data, it can still benefit by sharing these parameters with clients having similar data distribution. This observation emphasizes the importance of considering not only the sensitivity to non-IID data but also the similarity of data distribution when determining which parameters should be localized in PFL. This paper introduces a novel guideline for client collaboration in PFL. Unlike existing approaches that prohibit all collaboration of sensitive parameters, our guideline allows clients to share more parameters with others, leading to improved model performance. Additionally, we propose a new PFL method named FedCAC, which employs a quantitative metric to evaluate each parameter's sensitivity to non-IID data and carefully selects collaborators based on this evaluation. Experimental results demonstrate that FedCAC enables clients to share more parameters with others, resulting in superior performance compared to state-of-the-art methods, particularly in scenarios where clients have diverse distributions. The code is integrated into our FL training framework: \href{https://github.com/kxzxvbk/Fling}{https://github.com/kxzxvbk/Fling}.
\end{abstract}

\section{Introduction}


\begin{figure*}[tb]
	\centering
	\subfigure[]{
		\label{effect of noniid in training stage}
		\includegraphics[width=0.3\linewidth]{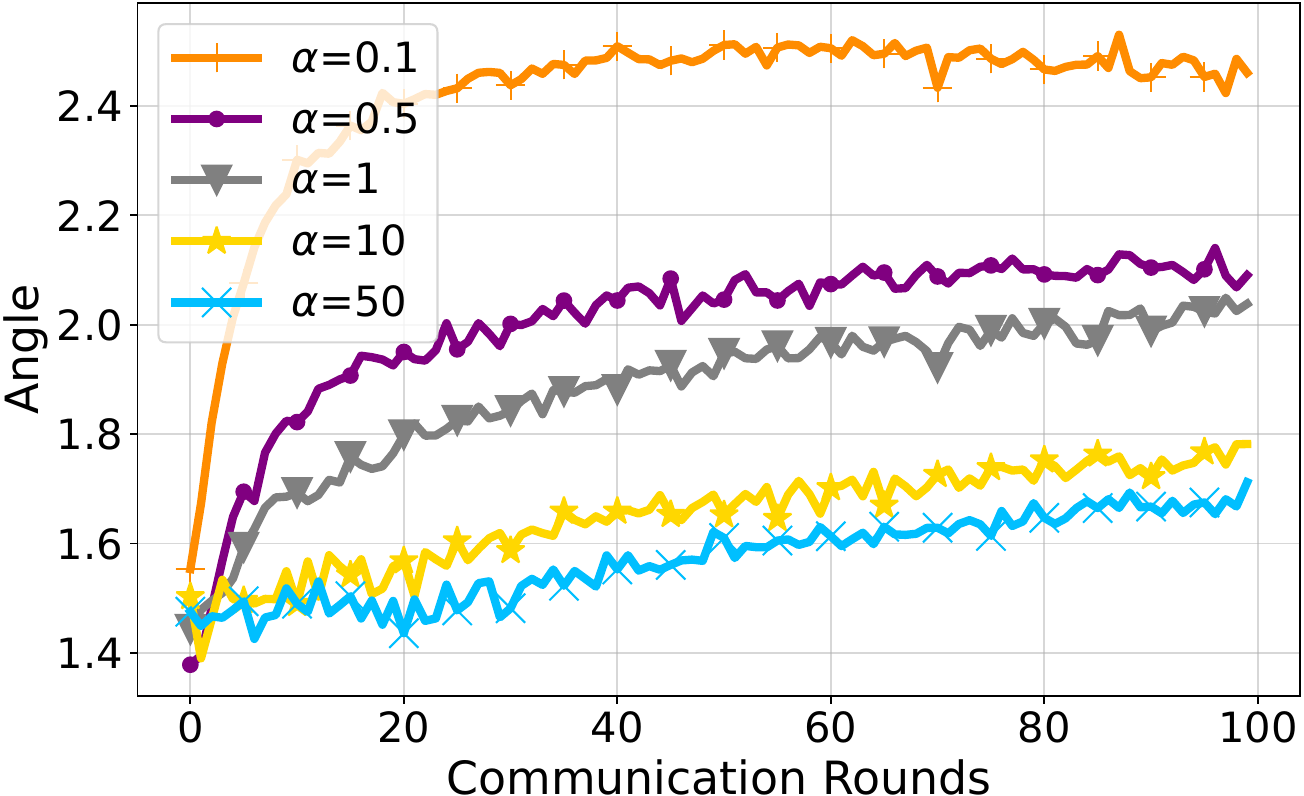}}
	\subfigure[]{
		\label{experimental example}
  
		\raisebox{0.3\height}{\includegraphics[width=0.68\linewidth]{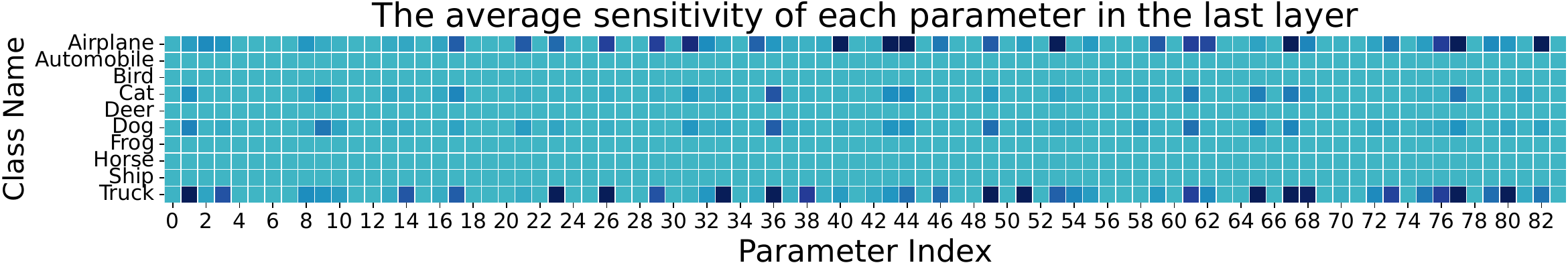}}}
	
	\caption{Two preliminary experiments. (a) verifies that the influence of non-IID varies with time. (b) verifies that parameters in the same layer have different sensitivities. }
	\label{different importance of parameters in one layer}
\end{figure*}

  
	

Federated Learning (FL) \cite{mcmahan2016federated} enables institutions or devices to train a global model collaboratively without exposing raw data. It has been applied in many scenarios, such as computer vision \cite{liu2020fedvision,doshi2022federated,wang2022atpfl}, finance \cite{long2020federated}, healthcare \cite{rieke2020future,sadilek2021privacy}, etc. However, when the data among clients are non-independent and identically distributed (non-IID), the local tasks of clients are different, which brings a great challenge for training a global model for all clients \cite{zhang2021federated,gong2021ensemble,li2021model}. Personalized federated learning (PFL) \cite{fallah2020personalized,t2020personalized,li2021ditto} allows each client to train a personalized model to focus on its data distribution. PFL aims to enable each client to get as much help as possible from other clients while minimizing the impact of non-IID data. A key question in PFL is to decide which parameters of a client should be localized to reduce the influence of non-IID or shared with others to get help.

To deal with this question, according to the characteristics of different neural network layers, the current mainstream work localizes layers sensitive to non-IID data while globally sharing the others. FedRep \cite{collins2021exploiting} and FedBN \cite{li2021fedbn}, for example, personalize classifier layers and batch normalization (BN) layers, respectively. The rationale behind these methods is localizing parameters easily influenced by non-IID data to reduce the negative effect of collaboration. 

However, although localizing sensitive layers can reduce the impact of non-IID data, these methods are too conservative for collaboration. The client can not fully utilize other clients' knowledge, and it is difficult to adapt to various complex non-IID scenarios. In addition, these methods are mostly empirical and heuristic and do not give a guideline for client collaboration.

We find that client collaboration depends not only on parameter sensitivity but also on variations in data distribution between clients. In non-IID scenarios, some clients still possess similar data distribution, allowing them to collaborate on sensitive parameters and benefit from each other. In this paper, we first propose a guideline for client parameter collaboration in PFL that takes both factors into account. It can be formulated as 
\begin{equation}\label{guideline}
    U = I(\Psi \cdot \Omega).
\end{equation}
In the above equation, $U \in \{0, 1\}^N$ and $U_i=1$ denotes that this parameter should not get contribution from client $i$. $N$ denotes the total number of clients. $I(\cdot)$ is an indicator function.  $\Psi \in [0,1]^N$, and $\Psi_i$ indicates the data distribution difference between the current client and client $i$. A larger $\Psi_i$ implies the two clients have less similar data distributions. $\Omega \in [0,1]$ indicates the current sensitivity of the parameter to non-IID data. The larger the $\Omega$, the more sensitive it is to non-IID data. This guideline offers a cautiously aggressive collaboration. Even sensitive parameters (i.e., $\Omega$ is large) can still collaborate with clients with similar data distribution (i.e., $\Psi_i$ is small), which enables a client can collaborate more with others while avoiding the negative effect of non-IID data. Current methods can also be incorporated under this guideline. For each client $i$, they set $\Psi_j=1, i \ne j$ and select parameters sensitive to non-IID data in a layer-wise manner to set their $\Omega=1$.

In practical use of the guideline, $\Omega$ is a critical factor that must be taken into account. Our research shows that two effects should be considered. First, $\Omega$ varies over time. In the early training stage, the impact of non-IID data is minimal, and parameters are less sensitive to non-IID data. As the training progresses, the impact of non-IID data becomes more pronounced, and parameters are more susceptible to its effects. To illustrate this, we conduct an experiment involving two clients in different non-IID scenarios and calculate the angle between their gradients in each round. As shown in Figure \ref{effect of noniid in training stage}, in all non-IID scenarios (i.e., with all $\alpha$ values), the angle between the two gradients increases as the training progresses, indicating that the two clients are increasingly negatively influenced by each other. In other words, the impact of non-IID data grows over time.

Second, we discover that the current approach of selecting sensitive parameters layer-wise is too coarse-grained. The $\Omega$ value of parameters within the same layer can be different. To demonstrate this, we perform an experiment on the CIFAR-10 dataset with the CNN network. The proportion of samples of different classes in the training data is \text{airplane}:\text{truck}:\text{cat}:\text{dog}=4:4:1:1. We calculate the sensitivity of each parameter in the last layer of classifier, and the results are displayed in Figure \ref{experimental example}. Each square represents the sensitivity of a parameter in the model's last layer, with darker colors indicating greater sensitivity. Each row in Figure \ref{experimental example} corresponds to parameters related to a specific class. The results reveal that the sensitivity of parameters in the same layer differs significantly. This indicates that we must select sensitive parameters at the parameter level to accomplish fine-grained collaboration. Additionally, this experiment demonstrates that the sensitivity of parameters is linked to data distribution. The parameters associated with the class that has more samples are more sensitive. This underscores the significance of considering differences in client data distribution (i.e., $\Psi$ in Eq.~\eqref{guideline} ) when collaborating with sensitive parameters.

  
	

Building on the aforementioned guideline and observation, we introduce a new PFL method called FedCAC. FedCAC utilizes a sensitivity-based quantitative metric to access each parameter in each round, identifying the parameters sensitive to non-IID data as critical parameters and the rest as non-critical. All clients collaborate to train non-critical parameters, while FedCAC implements a time-varying collaboration strategy for critical parameters. Initially, each client receives assistance from a greater number of clients. As training progresses, it gradually collaborates only with clients that have more similar data distribution. Our primary contributions are summarized as follows:
\begin{itemize}
    \item We propose a client collaboration guideline in PFL, which comprehensively considers the differences in data distribution between clients and the sensitivity of parameters to non-IID data. This framework enables clients to collaborate more aggressively while carefully avoiding the negative effect of non-IID data.
    \item Building upon the guideline, we propose a new PFL method that selects sensitive parameters in a parameter-wise manner and implements time-varying collaboration for them, allowing them to benefit fully from the assistance of other clients, thereby improving performance.
    \item The experimental results on CIFAR-10, CIFAR-100, and Tiny ImageNet prove that the performance of our method is significantly improved compared with the existing methods.
\end{itemize}

\section{Related Work}
PFL \cite{ijcai2022p311,ijcai2022p357,chen2022on,tan2022fedproto} is a powerful means to address the non-IID problem in FL. 
The primary purpose of PFL is to enable personalized models to benefit from inter-client collaboration while mitigating the impact of non-IID. 
Among these advanced PFL methods, parameter decoupling and personalized aggregation are two representative methods.

\textbf{Parameter Decoupling} localizes some layers that are sensitive to non-IID data while sharing others globally. 
Some current studies choose critical layers based on the characteristics of the neural network, such as classifiers \cite{arivazhagan2019federated, collins2021exploiting}, BN layers \cite{li2021fedbn} or Adapter layers \cite{pillutla2022federated}. Some works also attempt to identify critical layers automatically via advanced techniques in deep learning, such as reinforcement learning \cite{sun2021partialfed} or hypernetwork \cite{ma2022layer}. These methods can effectively mitigate the negative impact of non-IID by letting localized layers adapt to the local task. Nonetheless, in these methods, the localized layers neglect to collaborate with others, resulting in poor performance in complex non-IID scenarios.


\textbf{Personalized Aggregation} lets clients with similar data distribution collaborate to alleviate the impact of non-IID data. Based on the attention mechanism, FedAMP \cite{huang2021personalized} designs an attention-inducing function that allows similar models to aggregate with large weight. FedFomo \cite{zhang2020personalized} evaluates the degree of data distribution similarity through the model's validation set on the client side. Give better-performing models a larger aggregation weight. APPLE \cite{ijcai2022p301} learns a directed relational vector to guide personalized model aggregation. However, in practice, it is a challenge to find clients with similar data distribution to collaborate with, especially when the data distributions on clients vary dramatically.


Briefly, current PFL methods only consider the influence of partial factors on collaboration and lack a comprehensive guideline for collaboration, which cannot adapt to various complex non-IID scenarios.

\section{Method}

\subsection{Overview of FedCAC}
\begin{figure}[tb]
		\centerline{\includegraphics[width=\linewidth]{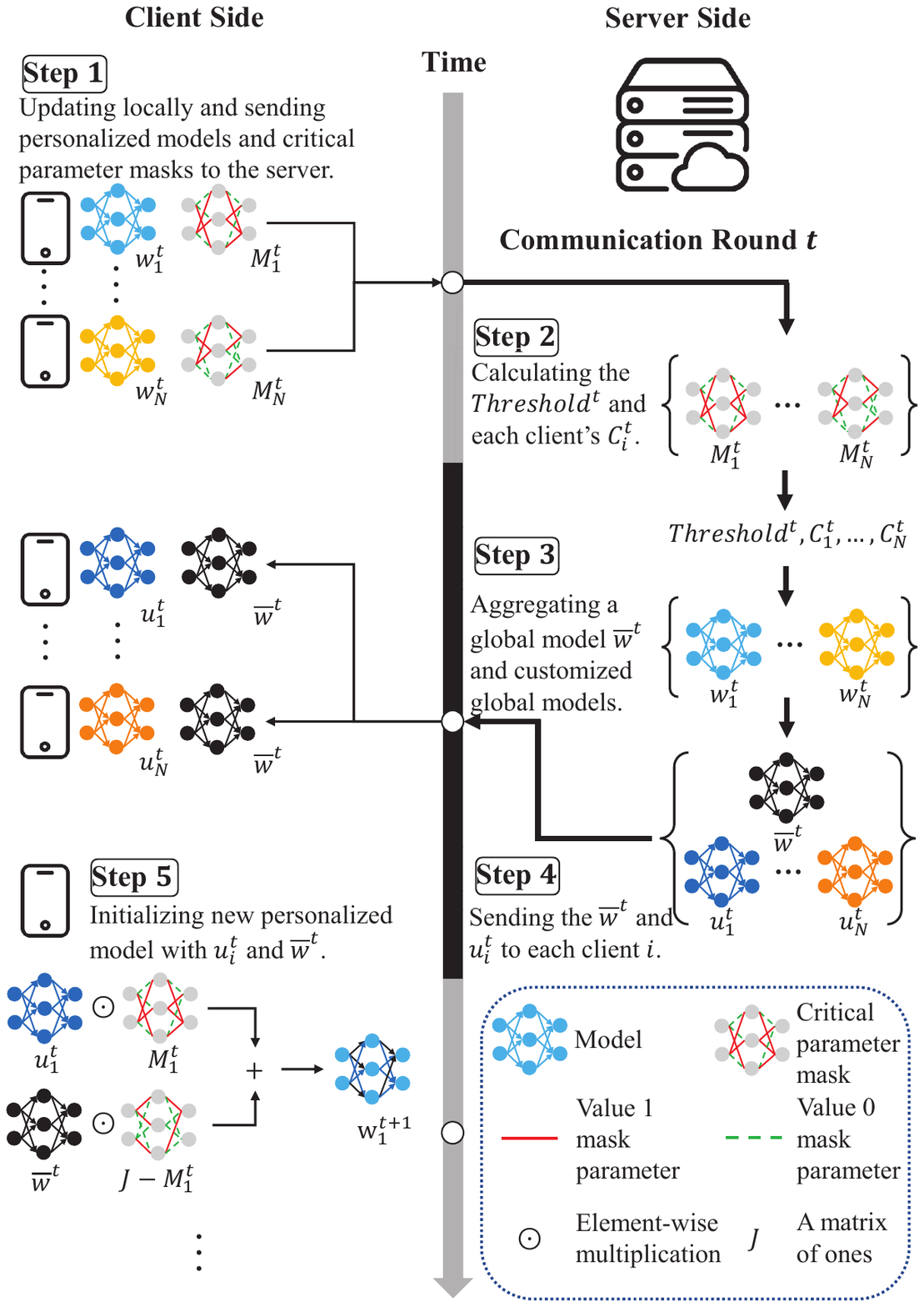}}
 \caption{The system workflow of FedCAC.}
		\label{overview of FedCAC}
\end{figure}

Before starting, we first give an overview of FedCAC. As shown in Figure \ref{overview of FedCAC}, the training process in each communication round of FedCAC can be summarized as follows:

\textbf{Step 1:} Each client $i$ trains the personalized model $w_i^t$ and selects critical parameters to generate a critical parameter mask $M_i^t$. Then, it sends the $w_i^t$ and $M_i^t$ to the server.

\textbf{Step 2:} The server calculates a threshold for critical parameter collaboration based on the overlap ratio of the locations of the critical parameters of different clients. Then, it selects a set of clients $C_i^t$ to collaborate critical parameters for each client $i$ based on the threshold.

\textbf{Step 3:} The server aggregates personalized models to generate a global model $\overline{w}^t$ and a customized global model $u_i^t$ for each client $i$ based on $C_i^t$.

\textbf{Step 4:} The server sends $\overline{w}^t$ and $u_i^t$ to each client $i$.

\textbf{Step 5:} Each client $i$ initializes the personalized model $w_{i}^{t+1}$ based on $\overline{w}^t$, $u_i^t$ and $M_i^t$.

In the following, we illustrate the details of FedCAC. Step 1 is introduced in section \ref{critical parameter selection}. Steps 2 to 5 are described in section \ref{critical parameter collaboration}.

\subsection{Problem Definition of PFL}\label{problem definition}
Traditional FL trains a global model for all clients, whereas PFL seeks to train a personalized model $w_i$ for each client $i$ that performs optimally on their respective local data distribution $D_i$. Let $L_i(w_i; D_i)$ denote the loss function of client $i$, then the training objective of PFL can be formulated as
\begin{linenomath}
\begin{equation}
    \min _{w_1,w_2,...,w_N} \sum_{i=1}^{N} L_i(w_i; D_i),
\end{equation}
\end{linenomath}
where $N$ is the total number of clients in FL. One approach to optimizing the objective described above is to utilize the local data on each client to minimize their corresponding loss function $L_i(w_i; D_i^{train})$, where the $D_i^{train}$ is the local data distribution of client $i$. However, clients participating in FL typically have very limited local data. As a result, the model can easily overfit to the $D_i^{train}$, resulting in poor generalization of $w_i$ on $D_i$. Collaborating with personalized models of other clients in the training process is essential. However, in non-IID scenarios, different clients have distinct data distributions (i.e., $D_i \ne D_j, i \ne j$), making collaboration with other clients potentially harmful to the performance of $w_i$ on their local tasks. Therefore, how to reduce the impact of non-IID when clients collaborate is a challenge in PFL.


\subsection{Sensitivity Based Parameter Selection}\label{critical parameter selection}

As discussed in Eq.~\eqref{guideline}, identifying parameters that are sensitive to non-IID (i.e., $\Omega$ in Eq.~\eqref{guideline}) is a critical aspect of collaboration design. $\Omega$ varies among different parameters and at different stages of training. We decouple these two factors in our approach. In this section, we begin by assessing the sensitivity of various parameters and selecting the most sensitive subset of parameters to non-IID data. In section \ref{critical parameter collaboration}, we examine the impact of the training stage when choosing collaborative clients for sensitive parameters.

In FedCAC, we employ a sensitivity-based approach for parameter selection. As discussed in previous research \cite{8953464,lee2018snip,zhang-etal-2022-pats}, the sensitivity of a parameter refers to the degree of variation in the model output or loss function when the parameter is set to 0. Specifically, given a model $w$ whose parameter set is expressed as $\Theta=\{\theta_1, ..., \theta_i,...,\theta_{n} \}$, then the sensitivity of the $i$-th parameter is defined as
\begin{linenomath}
\begin{equation}\label{def of sensitivity}
s_i=|L(\Theta)-L(\theta_1,...,\theta_{i-1},0,\theta_{i+1},...,\theta_{n})|,
\end{equation}
\end{linenomath}
where $L$ is the loss function. The sensitivity of a parameter is a useful metric for quantifying its significance to the local task. Furthermore, parameters crucial to the local task are more likely to degrade the model's accuracy when influenced by clients with different data distributions. As a result, the sensitivity of a parameter on the local task can also serve as a measure of the parameter's sensitivity to non-IID data. 

However, it can be seen from Eq.~\eqref{def of sensitivity} that evaluating each parameter requires an additional forward propagation, which significantly increases the computation cost. To solve this problem, we use the first-order Taylor approximation to substitute the calculation of $s_i$:
\begin{linenomath}
\begin{equation}\label{naive sensitivity}
\begin{split}
s_i &= |L(\Theta)-L(\theta_1,...,\theta_{i-1},0,\theta_{i+1},...,\theta_{n})| \\
&= |\nabla_{\theta_i}L(\Theta) \cdot \theta_i + R_1(\Theta)| \\
&\approx |\nabla_{\theta_i}L(\Theta) \cdot \theta_i|.
\end{split}
\end{equation}
\end{linenomath}
The gradients of all parameters can be obtained by one back propagation, which greatly reduces the computation overhead of $s_i$.

In FL, the client generally updates the model locally for several epochs. A simple way to apply Eq.~\eqref{naive sensitivity} to PFL is by computing sensitivity only in the final epoch. However, this approach may be prone to significant randomness. To overcome this limitation, we replace $\nabla_{\theta_i}L(\Theta)$ with the variation of model $\Delta \theta_i^t = \theta_i^{t,E} - \theta_i^{t,0}$, so as to calculate the sensitivity of each parameter to non-IID accurately. Here, the $\theta_i^{t,j}$ represents $\theta_i$ after the $j$-th ($j \in [1, E]$) local epoch in round $t$ ($t \in [1, T]$). $E$ is the maximum local epoch. $T$ is the maximum communication round. Based on this, our parameter sensitivity measurement metric can be expressed as
\begin{linenomath}
\begin{equation}\label{sensitivity index}
s_i^t= |\Delta \theta_i^t \cdot \theta_i^{t,E}|.
\end{equation}
\end{linenomath}

In each training round $t$ of FedCAC, each client $i \in [1, N]$ minimizes local loss function (e.g., cross-entropy loss in classification task) on local data by $E$ epochs to update personalized model $w_i^t$. Next, it computes the sensitivity of each parameter by Eq.~\eqref{sensitivity index} and constructs the parameter sensitivity matrix $S_i^t$. Based on $S_i^t$, each client selects the most sensitive parameters as \textit{critical} parameters, and the remaining parameters are considered \textit{non-critical}. We utilize a hyperparameter $\tau \in [0,1]$ to control the proportion of critical parameters to the total parameters. Because the distribution of parameter values in different neural network layers is different, critical parameters are selected layer by layer. That is, parameters with the top-$\tau$ sensitive in each layer are identified as critical parameters. We use a mask matrix $M_i^t$ to mark the critical parameters of client $i$. Each element
\begin{linenomath}
\begin{equation}\label{critical parameter mask}
m_{j,k}^t = 
\begin{cases}
1, & \text{if $S^t_{i,j,k}$ in the top-$\tau$ largest of $S^t_{i,j}$} \\
0, & \text{otherwise}
\end{cases}
\end{equation}
\end{linenomath}
in $M_i^t$ marks whether the $k$-th parameter of the $j$-th layer is a critical parameter. $S_{i,j}^t$ denotes the sensitivity set of parameters at the $j$-th layer of the model of client $i$. $S^t_{i,j,k}$ denotes the $k$-th item of $S_{i,j}^t$. 

It should be noted that by Eq.~\eqref{critical parameter mask}, the $M_i^t$ is a binary mask. It only needs 1 bit to store each element. The transmission cost of a mask is about 3\% compared to transmitting a model (32-bit per element). Therefore, the mask matrix introduced by FedCAC does not significantly increase the client's storage overhead and uplink communication cost.

\subsection{Parameter Collaboration in FedCAC}\label{critical parameter collaboration}
Once we decouple the critical and non-critical parameters, we proceed to design collaboration strategies for them on the server. According to Eq.~\eqref{guideline}, non-critical parameters' $\Omega$ are quite small, and they can safely collaborate with all clients. Conversely, critical parameters have a large $\Omega$ and need to collaborate with clients that have similar data distributions carefully.  Furthermore, as illustrated in Figure \ref{effect of noniid in training stage}, critical parameters can collaborate with more clients in the early training stage and gradually collaborate with clients whose data distributions are more similar. However, due to the privacy constraints in FL, we can not know the data distribution of clients. As shown in Appendix~\ref{appendix:similarity}, we find that the overlap ratio of client critical parameter locations can indicate the similarity degree of client data distribution, which can help us know the $\Psi$ in Eq.~\eqref{guideline}.

With the above idea, we design a time-varying collaboration strategy for critical parameters. We utilize $O_{i,j}^t= \frac{||M_i^t-M_j^t||_1}{2n}$ to represent the overlap ratio of critical parameter locations of the client $i$ and client $j$ in round $t$, where $||\cdot||_1$ denotes the $L_1$ norm and $n$ is the total number of parameters. In each training round $t$, after receiving $M_i^t$ from each client, the server calculates a threshold:
\begin{linenomath}
\begin{equation}\label{calculate threshold}
    \text{Threshold}^t = O_{avg}^t + \frac{t}{\beta} \times (O_{max}^t - O_{avg}^t),
\end{equation}
\end{linenomath}
where $O_{avg}^t=\frac{1}{N \cdot (N-1)} \sum_{i \ne j} O_{i,j}^t$ and $O_{max}^t=\max_{i\ne j}\{O_{i,j}^t \}$. The extent of assistance provided to the critical parameters by other clients is determined by $\beta \in [1, T]$. The higher the $\beta$, the more substantial the help provided to the critical parameters. Based on the $\text{Threshold}^t$, the server selects a set of collaboration clients for each client to collaborate critical parameters by
\begin{linenomath}
\begin{equation}\label{calculate C for each client}
    C_i^t = \{j| O_{i,j}^t \ge \text{Threshold}^t, j \ne i \}.
\end{equation}
\end{linenomath}
As the training progresses, the value of $\text{Threshold}^t$ gradually rises, resulting in a reduction in $|C_i^t|$. This implies that the client $i$ collaborates only with clients whose data distribution is more similar. Once $t > \beta$, each client trains critical parameters independently.


After obtaining the set of collaborative clients $C_i^t$, the server first aggregates a global model
\begin{linenomath}
\begin{equation}\label{global aggregating}
    \overline{w}^t = \frac{1}{N} \sum_{i=1}^{N} w_i^t
\end{equation}
\end{linenomath}
for all clients to collaborate on non-critical parameters. It also aggregates a customized global model 
\begin{linenomath}
\begin{equation}\label{intermediate aggregating}
    u_i^t = \frac{1}{|C_i^t|+1} \sum_{j \in C_i^t \cup \{i\}} w_j^{t},
\end{equation}
\end{linenomath}
for each client $i$ to collaborate on critical parameters. After receiving $\overline{w}^t$ and $u_i^t$ from the server, the client $i$ initializes the personalized model $w_i$ based on
\begin{linenomath}
\begin{equation}\label{local initialization}
    w_i^{t+1} = u_i^t \odot M_i^t + \overline{w}^t \odot (J-M_i^t),
\end{equation}
\end{linenomath}
where $J$ is a matrix with every element equal to one. In this way, FedCAC allows clients to aggressively get help from more clients, while carefully reducing the impact of non-IID data by taking into account parameter sensitivity, training stages, and differences in data distribution.


The details of the training process are summarized as pseudocode in Algorithm \ref{alg:FedCAC}.

\begin{algorithm}[htb]
	\caption{FedCAC}
	\label{alg:FedCAC}
	{\small
		\begin{algorithmic}
			\STATE {\bfseries Input:} Each client's initial personalized model $w_i^1$;
			Number of clients $N$;
			Total communication round $T$; Local epoch number $E$;
			Hyperparameters $\tau$, $\beta$.
			\STATE {\bfseries Output:} Personalized model $w_i^T$ for each client. \\
			\FOR {$t = 1$ to $T$}
            \STATE \textbf{Client-side:}
			\FOR{$i=1$ to $N$ \textbf{in parallel}}
            \STATE Updating $w_i^t$ for $E$ local epochs.
            \STATE Evaluating the sensitivity of each parameter by Eq~.\eqref{sensitivity index}.
            \STATE Obtaining critical parameter mask matrix $M_i^t$ by Eq.~\eqref{critical parameter mask}.
            \STATE Sending $w_i^t$ and $M_i^t$ to the server.
			\ENDFOR
            \STATE \textbf{Server-side:}
			\STATE Calculating $C_i^t$ for each client $i$ by Eq.~\eqref{calculate threshold} and Eq.~\eqref{calculate C for each client}.
            \STATE Aggregating a global model $\overline{w}^t$ by Eq.~\eqref{global aggregating}.
            \STATE Aggregating a $u_i^t$ for each client $i$ by Eq.~\eqref{intermediate aggregating}.
            \STATE Sending $\overline{w}^t$ and $u_i^t$ to each client $i$.
            \STATE \textbf{Client-side:}
            \FOR{$i=1$ to $N$ \textbf{in parallel}}
            \STATE Initializing $w_i^{t+1}$ with $\overline{w}^t$ and $u_i^t$ by Eq.~\eqref{local initialization}.
            \ENDFOR
			\ENDFOR
	\end{algorithmic}}
\end{algorithm}

\section{Experiments}

\subsection{Dataset Settings}
We conduct experiments on three datasets, including CIFAR-10 \cite{krizhevsky2010cifar}, CIFAR-100 \cite{krizhevsky2009learning}, and Tiny ImageNet \cite{le2015tiny}. Meanwhile, to verify the effectiveness of our method in different scenarios, we adopt two commonly used non-IID scenarios for experiments. One is pathological non-IID, and the other is Dirichlet non-IID.

\textbf{Pathological non-IID} is one of the most common non-IID settings in FL proposed by \cite{mcmahan2017communication}. In this setting, each client is randomly assigned two classes of data, and the data distributions among clients are very different.

\textbf{Dirichlet non-IID} is also a commonly used non-IID setting in FL \cite{hsu2019measuring,lin2020ensemble,kim2022multi,wu2022pfedgf}. In this setting, each client's data are drawn from the Dirichlet distribution, with $q \sim Dir(\alpha p)$, where $p$ is the prior distribution of all classes, and $\alpha$ is a hyperparameter used to control the degree of non-IID. As the $\alpha$ increases, the class imbalance degree of each client gradually decreases, resulting in more challenging local tasks (i.e., an increase in the number of classes and a decrease in the sample size for each class). At the same time, the difference in data distribution among clients gradually decreases while the variety of client data distribution increases. Therefore, Dirichlet non-IID is an effective way to evaluate the performance of methods in various and complex non-IID scenarios. To aid in intuitive understanding, we provide a visualization of the data partitioning in Appendix~\ref{appendix:dirichlet noniid}.

To highlight the effectiveness of the client collaboration, we assign a small amount of data to each client. Each client has 500 training samples and 100 test samples. Test data and training data have the same data distribution.

\subsection{Implementation Details}
We evaluate FedCAC against four SOTA methods, namely FedPer \cite{arivazhagan2019federated}, FedRep \cite{collins2021exploiting}, FedBN \cite{li2021fedbn}, and FedAMP \cite{huang2021personalized}. 
For the hyperparameters specific to all methods, we follow the optimal hyperparameter combination stated in their paper. In our methods, we adopt the SGD optimizer with a learning rate of 0.1. For federated learning training-related hyperparameters, we set the number of clients $N=40$, the number of local epochs $E=5$, and batch size equals 100. The experiments are conducted with the following settings: the maximum communication round T is set to 500 to ensure full convergence. In each run, we evaluate the uniform averaging test accuracy across all clients in each communication round and select the best accuracy as the final result. The ResNet \cite{he2016deep} network structure is used, specifically ResNet-8 for CIFAR-10, and ResNet-10 for CIFAR-100 and Tiny ImageNet. It should be noted that the BN layer is included in ResNet. The Statistical parameters \textit{running\_mean} and \textit{running\_var} in the BN layers cannot be used to calculate sensitivity by Eq.~\eqref{sensitivity index}. Following the work in \cite{li2021fedbn}, we set these parameters as critical parameters. Each experiment is repeated five times, and the mean and standard deviation are reported.


\subsection{Comparison with SOTA Methods}

\begin{table}[tb]
	\vskip 0in
	\begin{center}
		\begin{small}
				\begin{tabular}{lcccr}
					\toprule
					Methods & CIFAR-10 & CIFAR-100 & Tiny \\
					\midrule
					FedAvg & 54.33 $\pm$ 3.03 & 25.68 $\pm$ 1.31 & 17.93 $\pm$ 2.81 \\
					Separate & 85.85 $\pm$ 0.93 & 88.97 $\pm$ 1.10 & 86.43 $\pm$ 1.08 \\
					FedAMP & 88.88 $\pm$ 0.83 & 91.80 $\pm$ 0.61 & 88.23 $\pm$ 0.52 \\
					FedRep & 87.10 $\pm$ 0.91 & 90.25 $\pm$ 0.60 & 87.22 $\pm$ 0.94 \\
					FedBN & 87.02 $\pm$ 1.41 & 91.74 $\pm$ 0.62 & 89.41 $\pm$ 0.32 \\
					FedPer & 87.51 $\pm$ 0.95 & 90.33 $\pm$ 0.60 & 87.17 $\pm$ 1.10 \\
					\midrule
					FedCAC & \textbf{89.77 $\pm$ 1.14} & \textbf{93.05 $\pm$ 0.90} & \textbf{90.36 $\pm$ 0.75} \\
					\bottomrule
				\end{tabular}
		\end{small}
	\end{center}
 \caption{Comparison results under Pathological non-IID.}
 \label{pathological noniid}
	\vskip -0.0in
\end{table}
\textbf{Pathological non-IID.}
In the pathological non-IID scenario, each client is given only two classes of data and performs a simple binary classification task locally. From `Separate' in Table \ref{pathological noniid}, we can see that the client can obtain high accuracy by training itself.
However, data distribution among clients varies greatly, making collaboration a challenge. Direct collaboration with other clients can actually have a negative impact, resulting in poor performance for FedAvg. FedRep, FedBN, and FedPer perform better by personalizing the layers that are sensitive to non-IID data, thereby mitigating the negative impact. FedAMP selects clients with similar data distributions to collaborate, which reduces the impact of non-IID data and leads to better model performance. However, as the dataset difficulty increases (from CIFAR-10 with 10 classes to Tiny Imagenet with 200 classes), it becomes harder for clients to find similar clients for collaboration, which reduces FedAMP's advantage. Our approach, which allows sensitive parameters to receive help from clients with similar data distributions and insensitive parameters to receive help globally, overcomes this limitation and achieves the best performance across all datasets.

\begin{table*}[tb]
\footnotesize
\begin{center}
\begin{tabular}{@{}c|ccc|ccc|ccc@{}}
\toprule
       & \multicolumn{3}{c|}{CIFAR-10}              & \multicolumn{3}{c|}{CIFAR-100}              & \multicolumn{3}{c}{Tiny Imagenet}           \\ \midrule
Method & $\alpha=0.1$ & $\alpha=0.5$ & $\alpha=1.0$ & $\alpha=0.01$ & $\alpha=0.1$ & $\alpha=0.5$ & $\alpha=0.01$ & $\alpha=0.1$ & $\alpha=0.5$ \\ \midrule
FedAvg & 54.36$\pm$2.73 & 60.41$\pm$1.36 & 60.91$\pm$0.72 & 27.83$\pm$3.03 & 34.16$\pm$0.70 & 32.78$\pm$0.23 & 20.56$\pm$0.25 & 21.26$\pm$1.28 & 20.32$\pm$0.91 \\
Separate & 81.45$\pm$1.60 & 60.15$\pm$0.86 & 52.24$\pm$0.41 & 82.91$\pm$0.96 & 45.57$\pm$0.90 & 22.65$\pm$0.51 & 64.47$\pm$3.11 & 24.07$\pm$0.62 & \; 8.75$\pm$0.30 \\
FedAMP & 84.99$\pm$1.82 & 68.26$\pm$0.79 & 64.87$\pm$0.95 & 85.11$\pm$1.15 & 46.68$\pm$1.06 & 24.74$\pm$0.58 & 69.86$\pm$2.03 & 27.85$\pm$0.71 & 10.70$\pm$0.32 \\
FedRep & 84.59$\pm$1.58 & 67.69$\pm$0.86 & 60.52$\pm$0.72 & 84.40$\pm$1.43 & 51.25$\pm$1.37 & 26.97$\pm$0.33 & 70.45$\pm$1.11 & 30.83$\pm$1.05 & 12.14$\pm$0.28 \\
FedBN & 83.55$\pm$2.32 & 66.79$\pm$1.08 & 62.20$\pm$0.67 & 85.43$\pm$0.87 & 54.35$\pm$0.63 & 36.94$\pm$0.94 & 73.90$\pm$1.45 & 33.34$\pm$0.71 & 19.61$\pm$0.35 \\
    FedPer & 84.43$\pm$0.47 & 68.80$\pm$0.49 & 64.92$\pm$0.66 & 83.35$\pm$0.78 & 51.38$\pm$0.94 & 28.25$\pm$1.03 & 72.63$\pm$0.78 & 32.33$\pm$0.31 & 12.69$\pm$0.42 \\
\midrule
FedCAC & \textbf{86.82$\pm$1.18} & \textbf{69.83$\pm$0.46} & \textbf{65.39$\pm$0.51} & \textbf{86.58$\pm$0.85} & \textbf{57.22$\pm$1.52} & \textbf{38.64$\pm$0.63} & \textbf{75.76$\pm$1.21} & \textbf{40.19$\pm$1.20} & \textbf{23.70$\pm$0.28} \\
\bottomrule
\end{tabular}
\end{center}
\caption{Comparison results under Dirichlet non-IID on CIFAR-10, CIFAR-100, and Tiny Imagenet.}
\label{dirichlet noniid}
\end{table*}

\textbf{Dirichlet non-IID.}
We also evaluate our approach in the Dirichlet non-IID scenario. 
We conduct experiments with $\alpha$ values of $\{0.1, 0.5, 1.0\}$ for CIFAR-10 and $\{0.01, 0.1, 0.5\}$ for CIFAR-100 and Tiny Imagenet.

Based on the experimental results shown in Table \ref{dirichlet noniid}, it can be observed that all SOTA methods achieve significant performance improvements when compared to the `Separate' approach in the Dirichlet non-IID scenario. Among these methods, FedAMP and FedPer show more obvious superiority on CIFAR-10. This can be attributed to the fact that it is easier for clients to find and collaborate with other clients having similar data distributions in the 10-classification dataset, thereby resulting in better model performance for FedAMP. Moreover, the 10-classification task also makes the classifier training simple, allowing the client to train the classifier effectively while reducing the influence of other clients, resulting in the better model performance of FedPer.

The experimental results on CIFAR-100 and Tiny Imagenet datasets show that the performance of SOTA methods differs significantly from that on CIFAR-10. With the increase in the number of classes, the variety of client data distribution increases, making it harder for FedAMP to find clients with similar data distributions, resulting in its models hardly getting help from others. The increase in the number of classes can also make local tasks harder, causing difficulty for clients to train their classifiers and increasing the risk of overfitting. Therefore, the superiority of FedPer reduces. 
Interestingly, most SOTA methods perform poorly when $\alpha=0.5$. Moreover, no SOTA method is better than FedAvg on the Tiny Imagenet dataset. This suggests that the current PFL approach does not fully leverage the knowledge of other clients when collaborating, leading to poor performance in the face of difficult datasets and complex client-side data distribution scenarios.

FedCAC can obtain the best model performance under all settings. It performs particularly well when the clients' data distributions are complex and their local tasks are difficult (e.g., $\alpha=0.1$ or $\alpha= 0.5$ in Tiny Imagenet). This demonstrates FedCAC can adapt to a variety of complex non-IID scenarios and achieve more efficient collaboration
by fully considering the client data distribution differences, training stages, and parameter sensitivity. Moreover, it confirms the effectiveness of the guideline proposed in Eq.~\eqref{guideline}.

\subsection{Component Analysis}\label{The Effect of Different Modules in FedCAC}
In this section, we verify the effect of various components introduced to FedCAC. 

	

\begin{figure}[tb]
	\centering
	\subfigure[CIFAR-10]{
		\label{Effect of sensitivity on CIFAR-10}
		\includegraphics[width=0.31\linewidth]{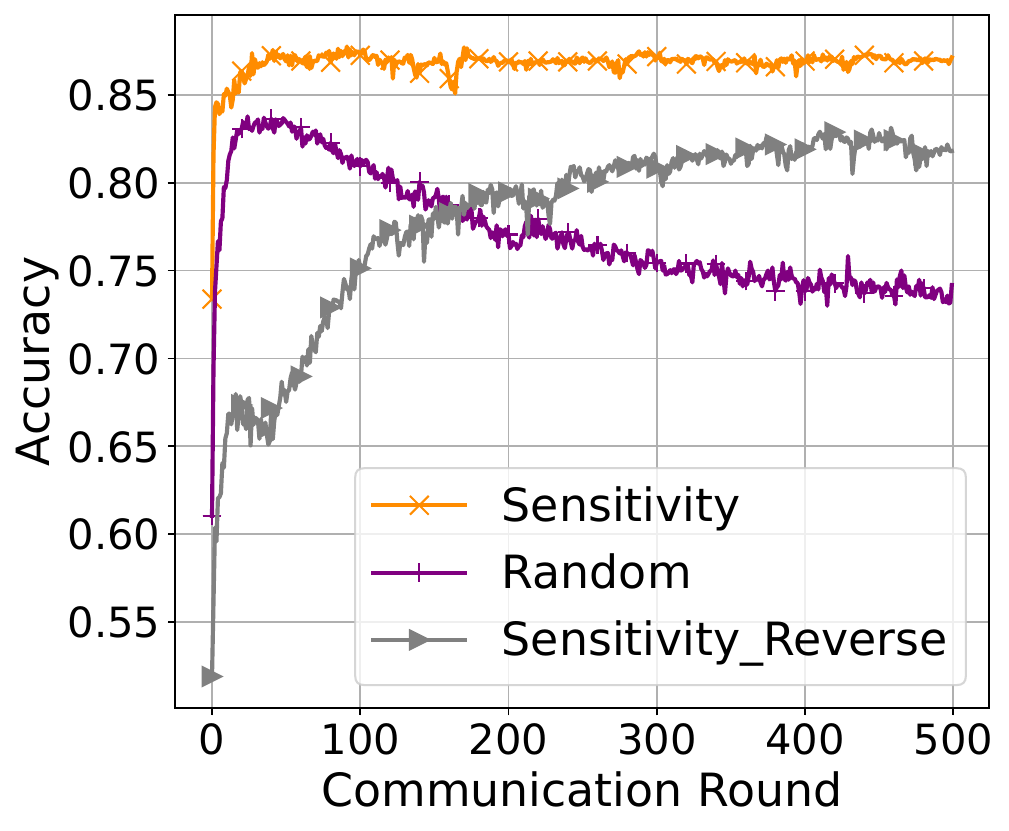}}
	\subfigure[CIFAR-100]{
		\label{Effect of sensitivity on CIFAR-100}
		\includegraphics[width=0.31\linewidth]{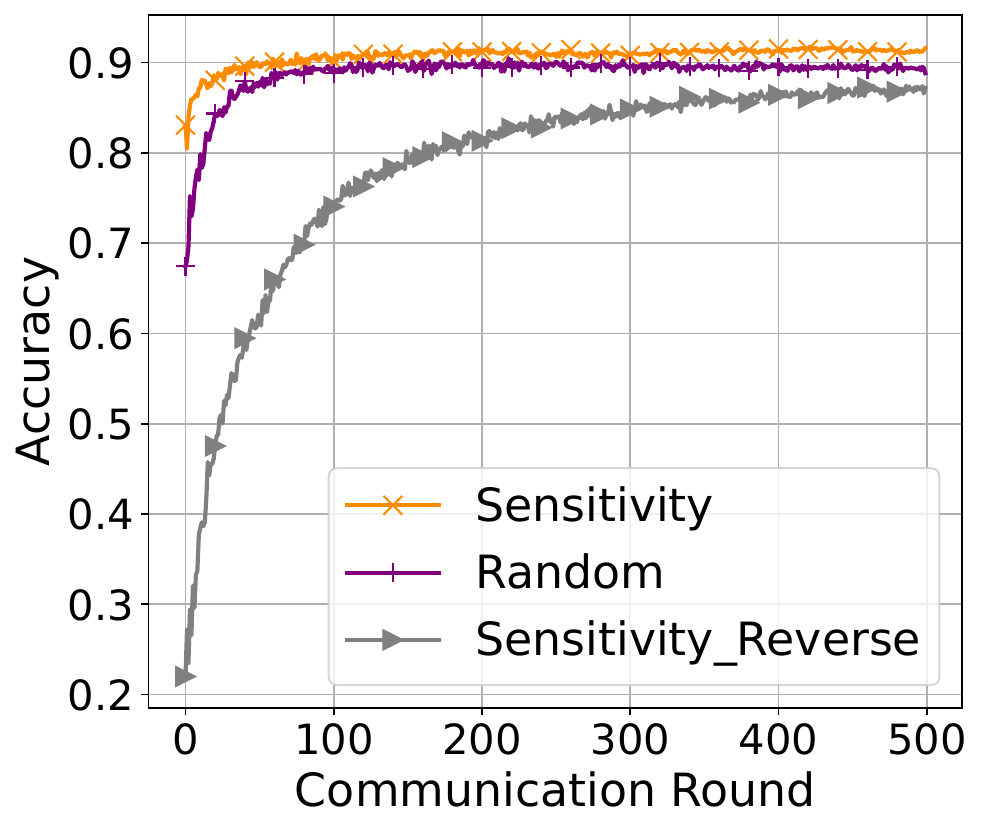}}
	\subfigure[Tiny Imagenet]{
		\label{Effect of sensitivity on TinyImagenet}
		\includegraphics[width=0.31\linewidth]{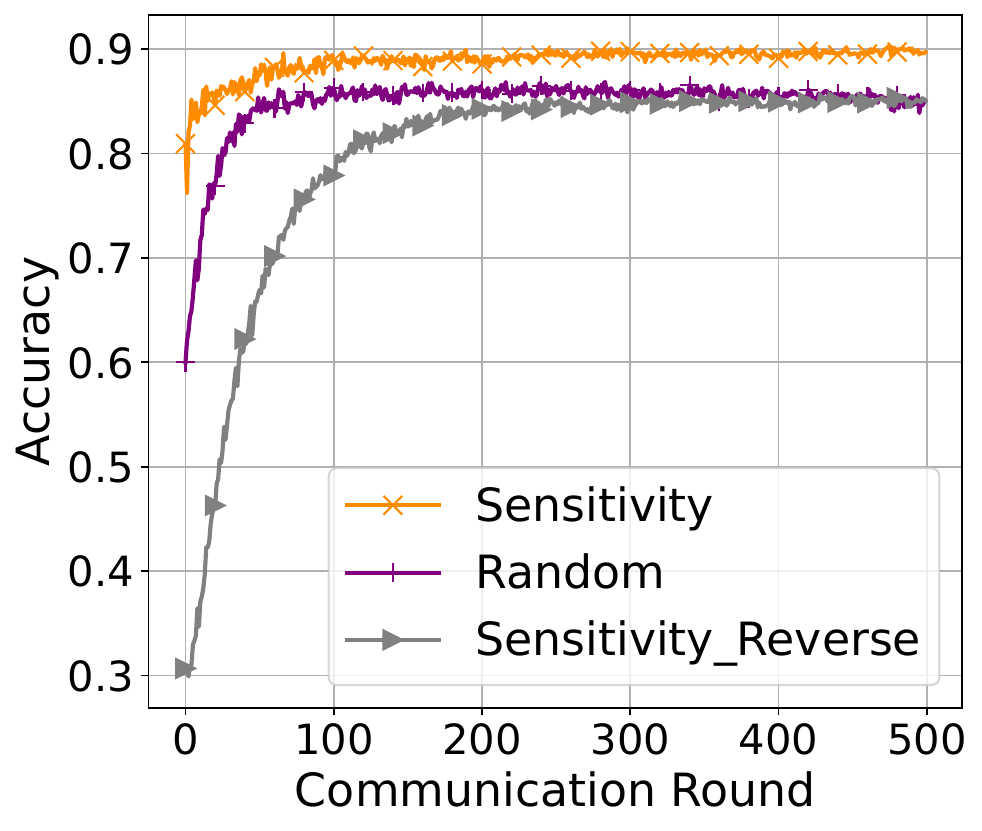}}
	
	\caption{Effect of different critical parameter selection schemes on CIFAR-10, CIFAR-100, and Tiny Imagenet datasets.}
	\label{Effect of sensitivity}
\end{figure}

\textbf{Effect of Critical Parameter Selection Schemes.}
We compare three critical parameter selection methods: `Sensitivity', which is based on sensitivity, `Random', which randomly selects some parameters as critical parameters in each round, and `Sensitivity\_Reverse', which uses our sensitivity-based method to choose insensitive parameters as the critical parameters. We conduct experiments on Pathological non-IID with hyperparameter $\tau=0.5$. During training, we localize all critical parameters to exclude the influence of critical parameter collaboration.

The results are shown in Figure \ref{Effect of sensitivity}. We can see that the accuracy of `Random' increases fast in the early stage, which indicates that it randomly selects some sensitive parameters and reduces the influence of non-IID data. However, its accuracy decreases as the training goes on, indicating that the influence of non-IID data is gradually increasing. The `Sensitivity\_Reverse' method results in slow convergence and poor final accuracy due to the client having a greater influence on the local task when collaborating. The `Sensitivity' method is superior to other methods in all datasets, in terms of convergence speed and accuracy, which indicates that our method can efficiently and accurately select parameters sensitive to non-IID data.

\begin{figure}[tb]
	\centering
	\subfigure[$\alpha=0.01$]{
		\label{The effect of critical parameter collaboration when alpha=0.01}
		\includegraphics[width=0.31\linewidth]{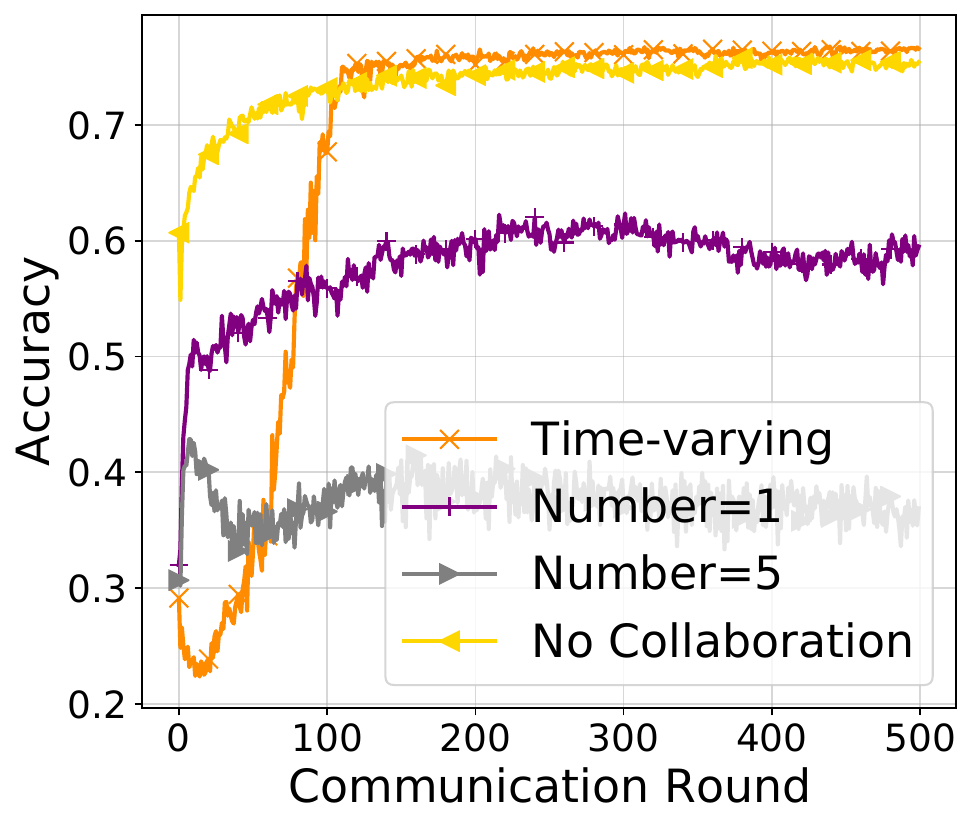}}
	\subfigure[$\alpha=0.1$]{
		\label{The effect of critical parameter collaboration when alpha=0.1}
		\includegraphics[width=0.318\linewidth]{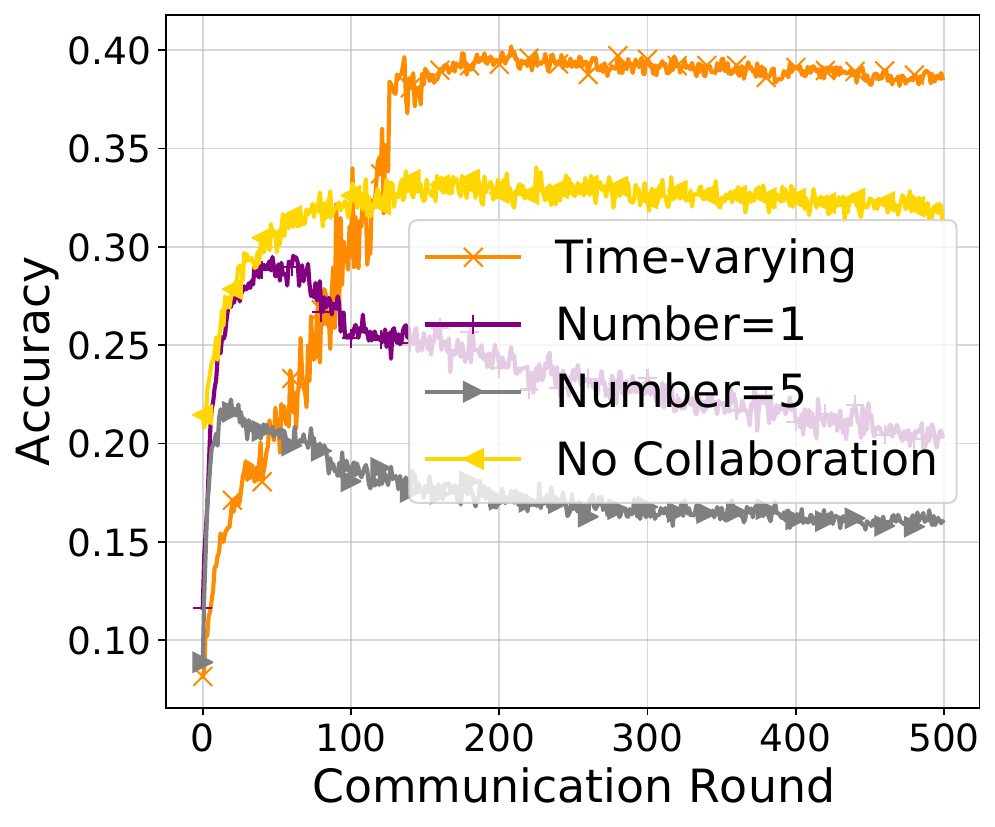}}
	\subfigure[$\alpha=0.5$]{
		\label{The effect of critical parameter collaboration when alpha=0.5}
		\includegraphics[width=0.318\linewidth]{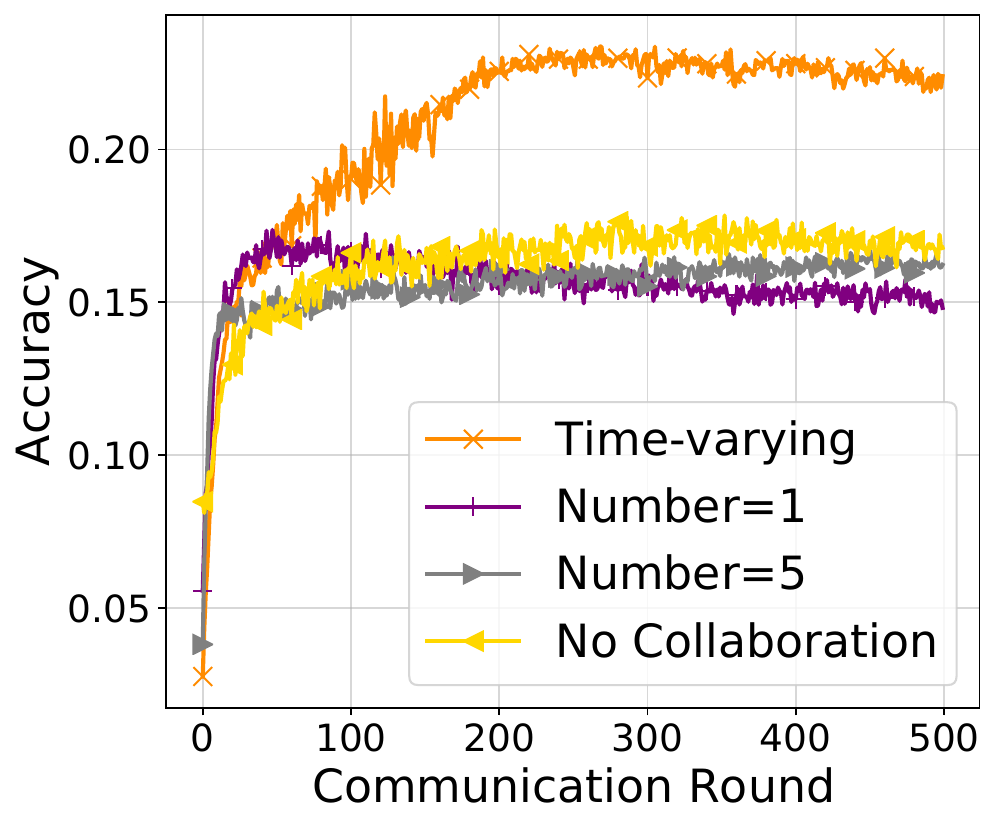}}
	
	\caption{Effect of different critical parameter collaboration schemes on Tiny Imagenet.}
	\label{Effect of collaboration on critical}
\end{figure}
\textbf{Effect of Critical Parameter Collaboration Schemes.}
To demonstrate the effectiveness and necessity of our critical parameter collaboration approach in various scenarios, we conduct experiments on the challenging Tiny Imagenet dataset under the Dirichlet non-IID setting, where we sample $\alpha \in \{0.01, 0.1, 0.5 \}$. We introduce two additional methods for comparison, aiming at highlighting the challenges involved in designing collaborative approaches for critical parameters. `Number=1' and `Number=5' represent methods where each client collaborates with the one and five clients with the highest overlap rate of critical parameter locations during the training process. As a baseline, we include the `No Collaboration' method, which localizes the client's critical parameters.

The experimental results are shown in Figure \ref{Effect of collaboration on critical}. When $\alpha=0.01$, there are only a few classes of data on each client, and the local task is relatively simple. The client itself training critical parameters can achieve better results. We can see that `No Collaboration' works well. In this case, the non-IID degree is high, and client-side collaboration can have a negative impact. We can see that `Number=1' and `Number=5' have a significant performance degradation compared to `No Collaboration'. And the more clients collaborate, the more obvious the decline. Although the `Time-varying' designed by us encourages clients to collaborate in the early training stage, resulting in a lower convergence rate than `No Collaboration', the critical parameters of the clients can also provide positive help to each other in the end, thus achieving better performance. Meanwhile, we can see that with the increase of $\alpha$, the local task of the client gradually becomes more difficult, and the client itself cannot train the critical parameters well. At this time, the client is more inclined to seek help from other clients, and the benefits of collaboration are also increasing. Compared with `No Collaboration', our `Time-varying' has increased performance more and more.

The above experiments illustrate that the collaboration of critical parameters is a huge challenge. It is not only necessary to consider the client data distribution similarity but also to consider the influence of non-IID in different training stages. This also shows that based on our guidelines, our approach can select the right collaboration clients at the right time, further improving the model's performance.

\begin{table}[tb]
	\vskip 0in
	\begin{center}
		\begin{small}
			\begin{sc}
				\begin{tabular}{lcccr}
					\toprule
					Methods & CIFAR-10 & CIFAR-100 & Tiny \\
					\midrule
					All & 86.82$\pm$1.18 & 57.22$\pm$1.52 & 40.19$\pm$1.20 \\
					As Critical & 85.42$\pm$1.39 & 55.15$\pm$1.10 & 37.97$\pm$0.63 \\
					\bottomrule
				\end{tabular}
			\end{sc}
		\end{small}
	\end{center}
 \caption{The effect of non-critical parameters collaboration on three datasets.}
 \label{effect of non-critical collaboration}
	\vskip -0.0in
\end{table}
\textbf{Effect of Collaboration on Non-critical Parameters.}
To validate the idea that non-critical parameters, which are less sensitive to non-IID, can benefit from collaborating with all clients, we conduct experiments in the Dirichlet non-IID scenario with $\alpha=0.1$.

The experimental results in Table~\ref{effect of non-critical collaboration} demonstrate that collaborating on non-critical parameters with all clients (denoted by `All') yields better performance than treating non-critical parameters like critical parameters (denoted by `As Critical') on all three datasets. This suggests that collaborating on non-critical parameters with all clients can provide more benefits than only collaborating with similar data distributions. These results support our guideline that non-critical parameters, which are less sensitive to non-IID, can collaborate with as many clients as possible for more help.

\subsection{Effect of Hyperparameters in FedCAC}
\begin{figure}[tb]
	\centering
	\subfigure[CIFAR-10]{
		\label{Effect of tau on CIFAR-10}
		\includegraphics[width=0.31\linewidth]{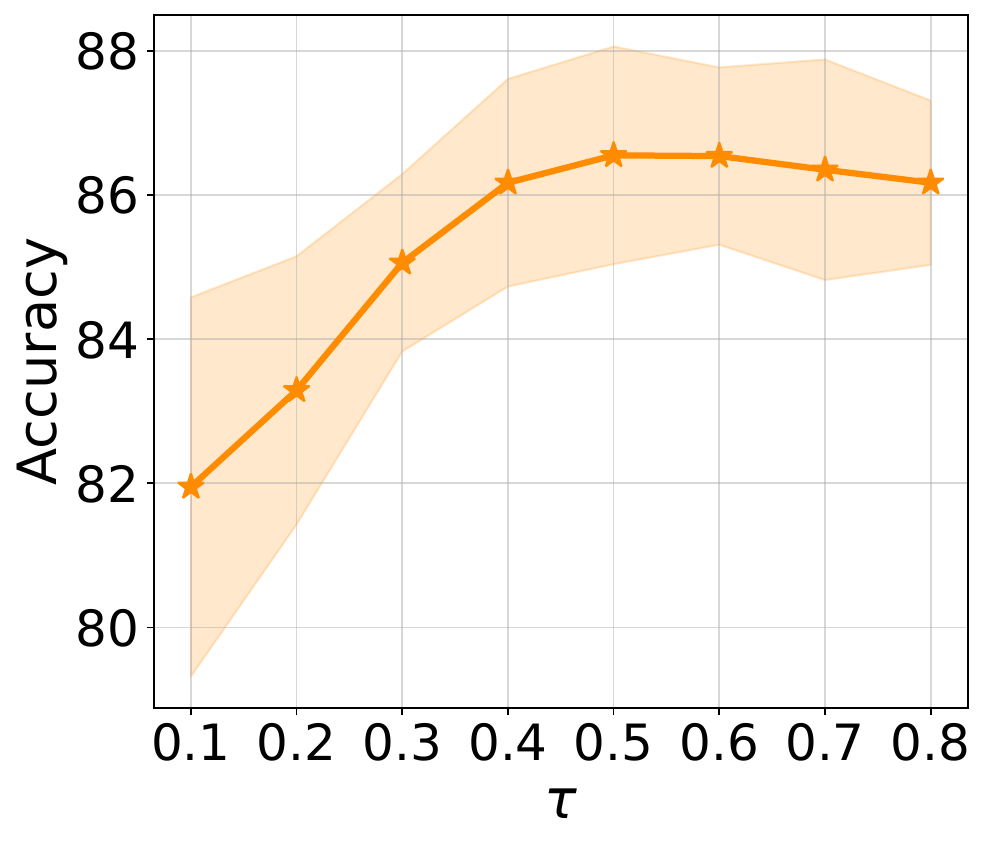}}
	\subfigure[CIFAR-100]{
		\label{Effect of tau on CIFAR-100}
		\includegraphics[width=0.31\linewidth]{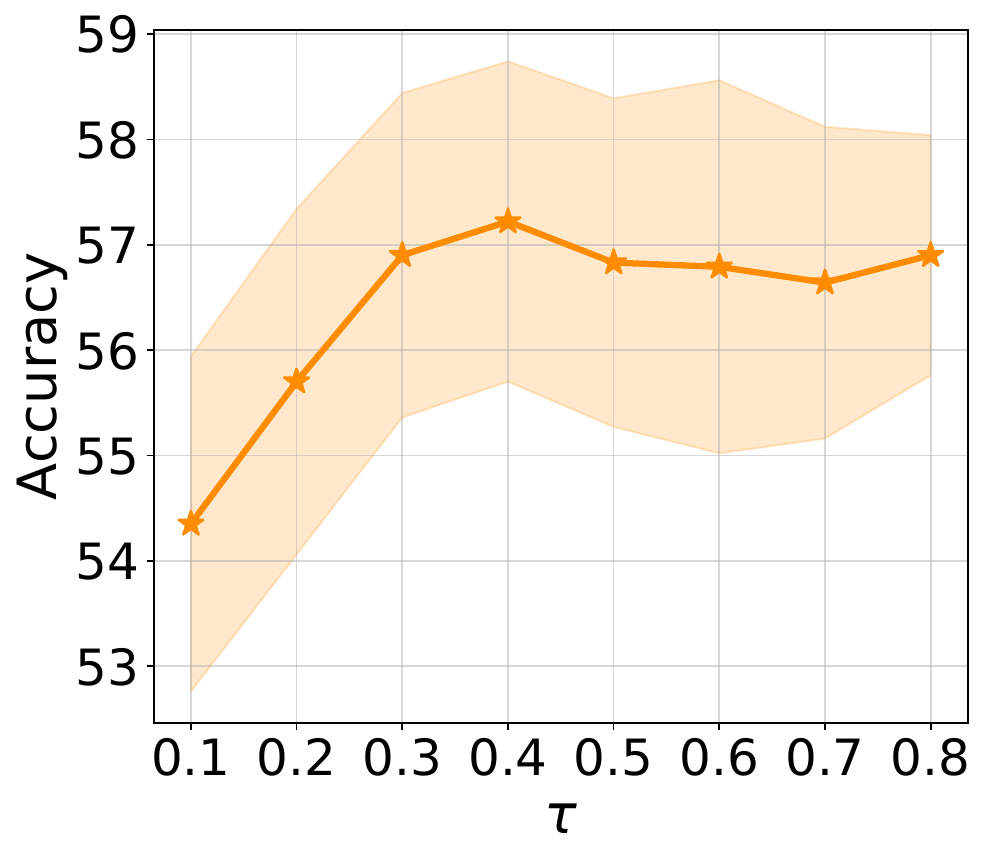}}
	\subfigure[Tiny Imagenet]{
		\label{Effect of tau on TinyImagenet}
		\includegraphics[width=0.316\linewidth]{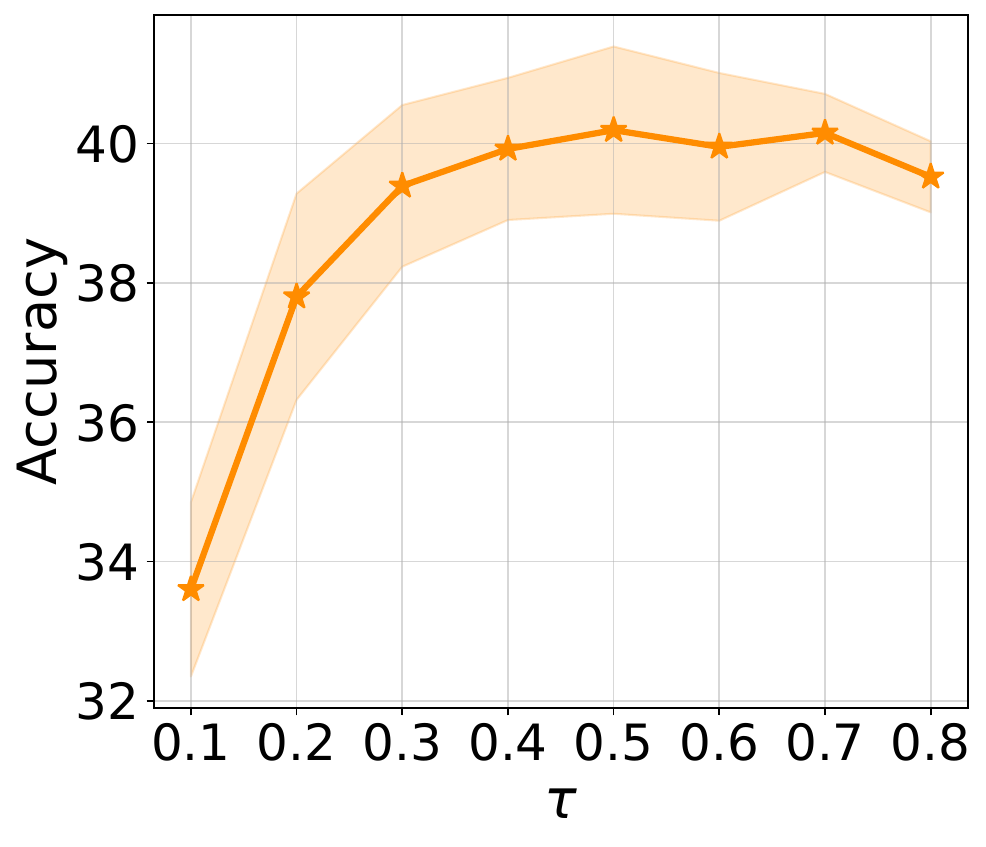}}
	
	\caption{Effect of $\tau$ in Dirichlet non-IID scenario with $\alpha = 0.1$.}
	\label{Effect of tau}
\end{figure}
\textbf{Effect of $\tau$.}
$\tau$ is a hyperparameter used to control the proportion of critical parameters. If $\tau$ is too small, some sensitive parameters may not be selected, and the collaboration among clients will be affected by non-IID, whereas a large value of $\tau$ may lead to a lack of help for insensitive parameters. We conduct the experiment in Dirichlet non-IID scenario with $\alpha=0.1$ to evaluate this intuition, and the experimental results are shown in Figure \ref{Effect of tau}. We observe that when the $\tau$ is small, the accuracy is very low, indicating that client collaboration significantly impacts each other's local task accuracy. As $\tau$ increases, the influence among clients gradually decreases, and the accuracy gradually increases. The accuracy is highest when $\tau$ equals around 0.5. When $\tau$ is increased again, the help clients get from other clients decreases, and the accuracy gradually decreases. This is in line with our expectations. Furthermore, we find that near the optimal $\tau$, the accuracy does not change much when the $\tau$ is changed slightly, indicating that FedCAC is robust to $\tau$. For practical use, we preferentially select $\tau=0.5$ and slightly adjust on this basis in different non-IID scenarios.

\begin{figure}[tb]
	\centering
	\subfigure[CIFAR-10]{
		\label{Effect of beta on CIFAR-10}
		\includegraphics[width=0.31\linewidth]{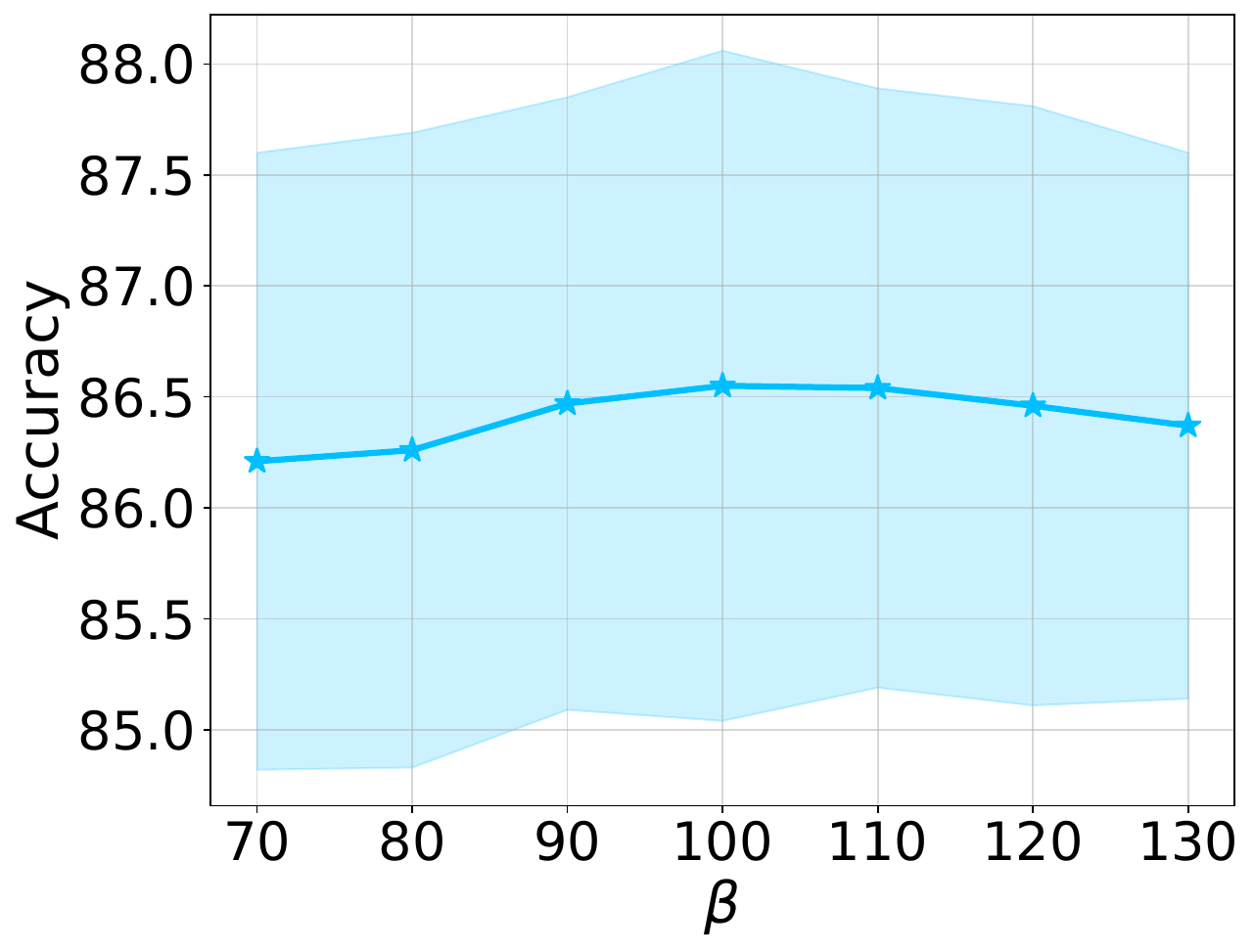}}
	\subfigure[CIFAR-100]{
		\label{Effect of beta on CIFAR-100}
		\includegraphics[width=0.31\linewidth]{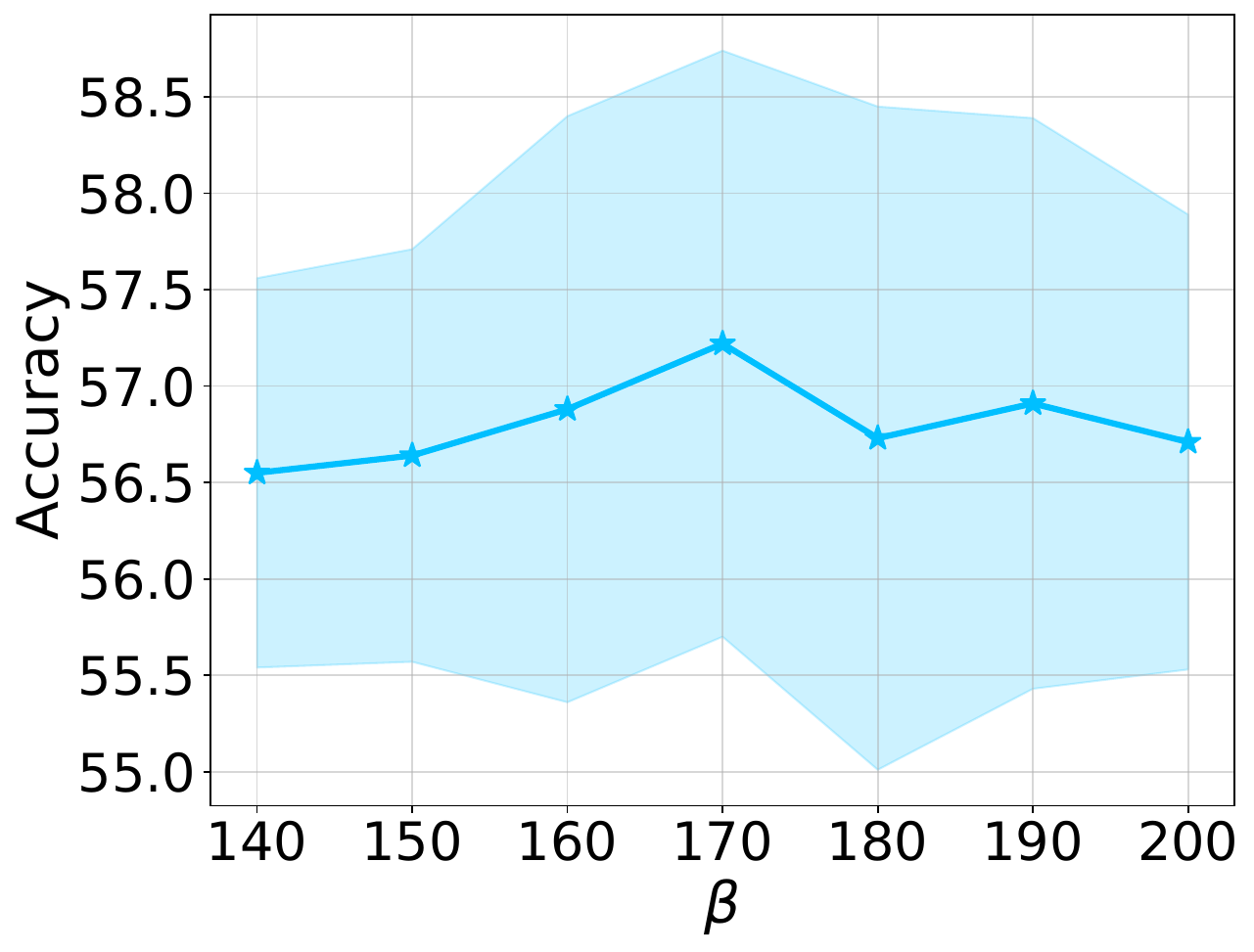}}
	\subfigure[Tiny Imagenet]{
		\label{Effect of beta on TinyImagenet}
		\includegraphics[width=0.316\linewidth]{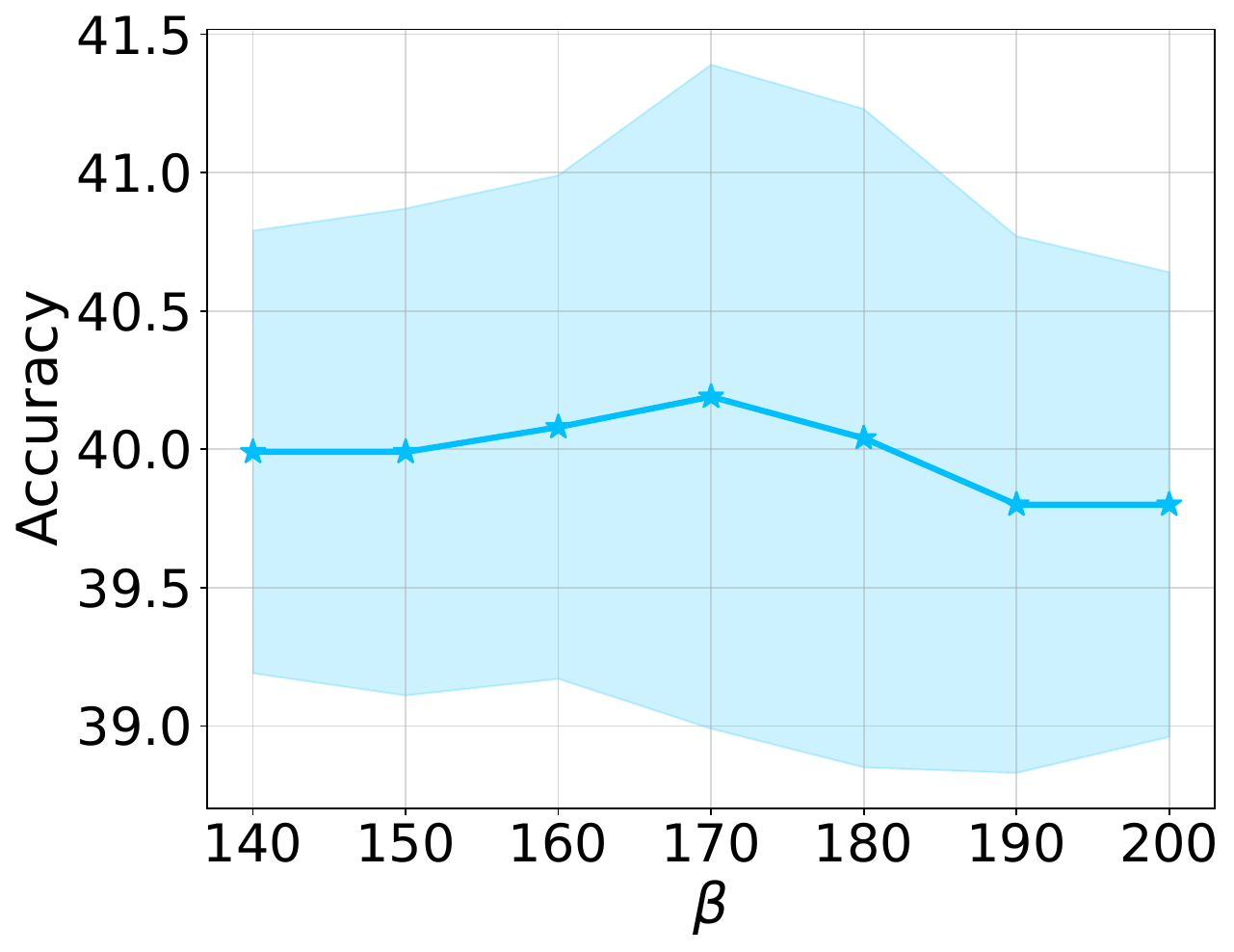}}
	
	\caption{Effect of $\beta$ in Dirichlet non-IID scenario with $\alpha = 0.1$.}
	\label{Effect of beta}
\end{figure}

\textbf{Effect of $\beta$.}
The hyperparameter $\beta$ controls the degree to which critical parameters of clients receive help from other clients. To evaluate the impact of $\beta$, we conduct the experiment in Dirichlet non-IID scenario with $\alpha=0.1$, and the experimental results are shown in Figure \ref{Effect of beta}. A low value of $\beta$ means that clients receive little help from others, leading to low accuracy. On the other hand, when $\beta$ is high, clients cannot timely reduce the number of collaborating clients, resulting in a greater influence of non-IID. It is observed that the accuracy is less sensitive to slight variations in $\beta$ near its optimal value, indicating that FedCAC is robust to $\beta$. In this paper, for the experiments on CIFAR-10, CIFAR-100, and Tiny Imagenet, we preferentially select $\beta=100$, $\beta=170$, and $\beta=170$, respectively, and make minor adjustments on this basis in different non-IID scenarios.

\section{Conclusion}
In this paper, we propose a comprehensive guideline for PFL collaboration that takes into account both the data distribution difference between clients and the sensitivity of each parameter.  Building on this guideline, we introduce a novel PFL method called FedCAC, which leverages a quantitative metric to evaluate the sensitivity of each parameter and dynamically selects clients with similar data distribution for the sensitive parameters to collaborate.  Our experimental results demonstrate that guided by the proposed guideline, our FedCAC method effectively makes each client obtain more help from others, resulting in superior performance in various complex non-IID scenarios.

\section*{Acknowledgments}
This work was supported in part by the National Natural Science Foundation of China under Grant No. 61976012.

{\small
\bibliographystyle{ieee_fullname}
\bibliography{egpaper_for_review}
}


\appendix
\section{Relationship Between Client Data Distribution Similarity and Location Overlap Ratio of Critical Parameter}\label{appendix:similarity}
\begin{figure}[tb]
\setlength{\abovecaptionskip}{0.cm}
\setlength{\belowcaptionskip}{-0.cm}
	\centering
	\subfigure[]{
		\label{Overlap matrix}
		\includegraphics[width=0.42\linewidth]{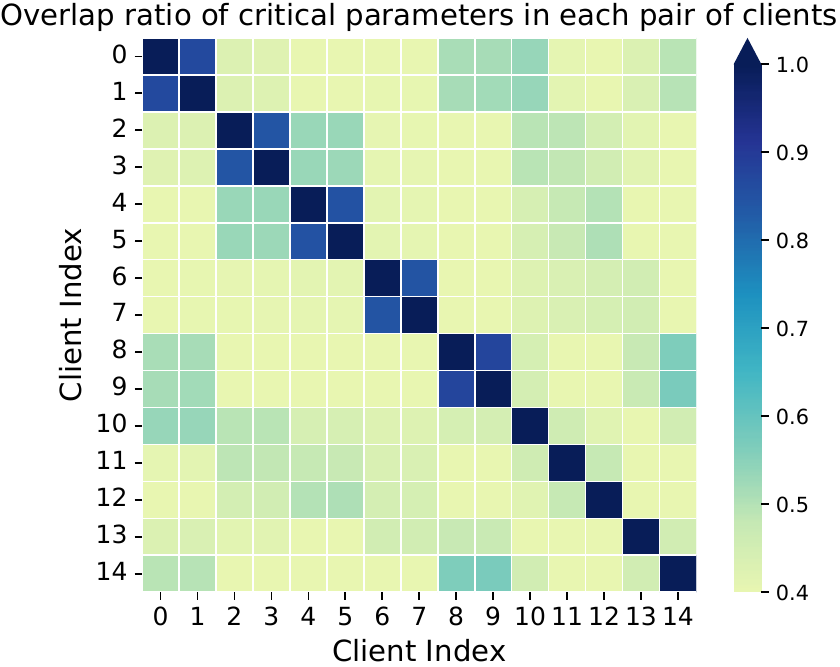}}
	\subfigure[]{
		\label{Visualize data distribution for 15 clients}
  
		\includegraphics[width=0.54\linewidth]{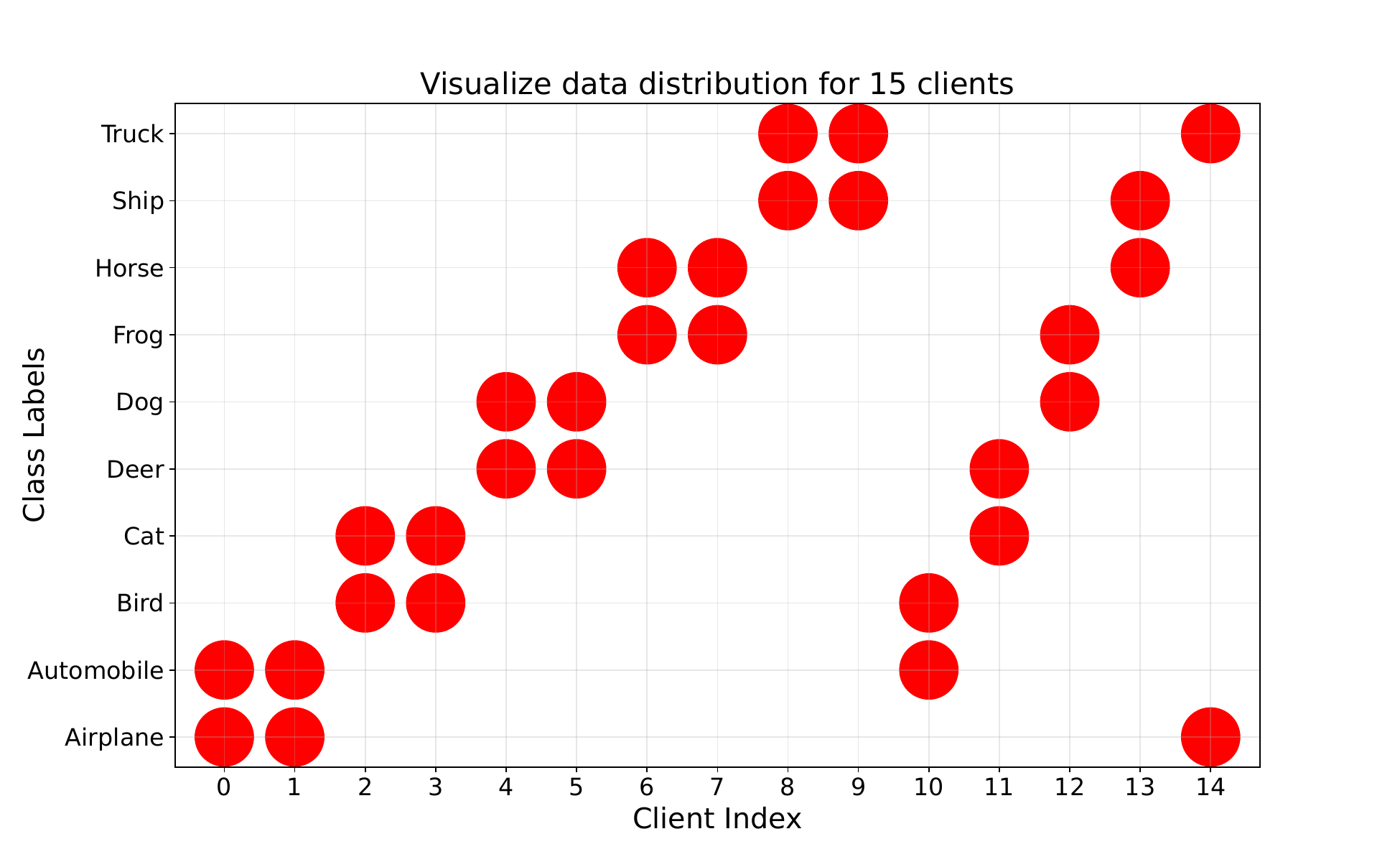}}
	
	\caption{An experiment to illustrate that clients with similar data distribution have similar locations of critical parameters.}
	\label{critical parameter location overlap}
\end{figure}
As we describe in section 1, a key factor affecting client parameter collaboration is the data distribution difference between clients (i.e., $\Psi$ in Eq.(1)). However, due to privacy constraints in FL, we can not know the data distribution of clients, which brings a challenge when implementing Eq.(1). In Figure 1(b) of section 1, we find that the sensitivity of parameters is related to the data distribution. Therefore, it is intuitive that clients with similar data distributions should have similar locations of their critical parameters. To demonstrate this intuition, as shown in Figure \ref{critical parameter location overlap}, we conduct an experiment with 15 clients. Figure \ref{Overlap matrix} shows the overlap ratio of the locations of critical parameters for any two clients. Figure \ref{Visualize data distribution for 15 clients} shows the data distribution for 15 clients. Take client 0 as an example. It has the same data distribution as client 1 and class overlap with clients 10 and 14, so their critical parameter location overlap ratio is high. Also, client 0 has a high overlap ratio with clients 8 and 9 due to the similarity of ‘Automobile’ and ‘Truck’ data. The result of the experiment is consistent with our intuition. Therefore, in our proposed FedCAC, we utilize the overlap ratio of critical parameter locations to indirectly reflect the client data distribution similarity.

\section{Visualization of data partitioning in Dirichlet non-IID scenarios}\label{appendix:dirichlet noniid}
\begin{figure}[tb]
\setlength{\abovecaptionskip}{0.cm}
\setlength{\belowcaptionskip}{-0.cm}
	\centering
	\subfigure[$\alpha=0.01$]{
		\includegraphics[width=0.45\linewidth]{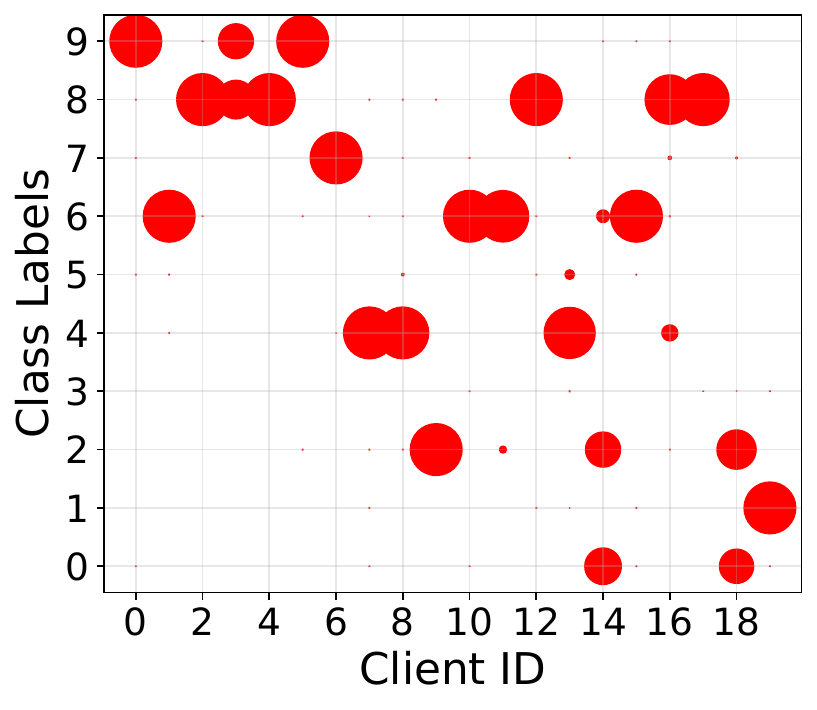}}
	\subfigure[$\alpha=0.1$]{
		\includegraphics[width=0.45\linewidth]{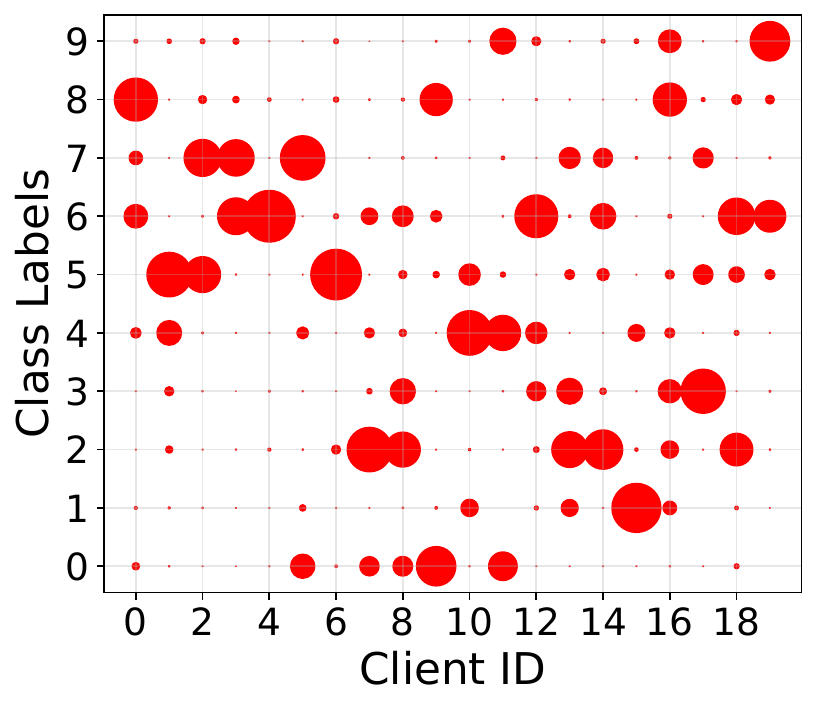}}
	\subfigure[$\alpha=0.5$]{
		\includegraphics[width=0.45\linewidth]{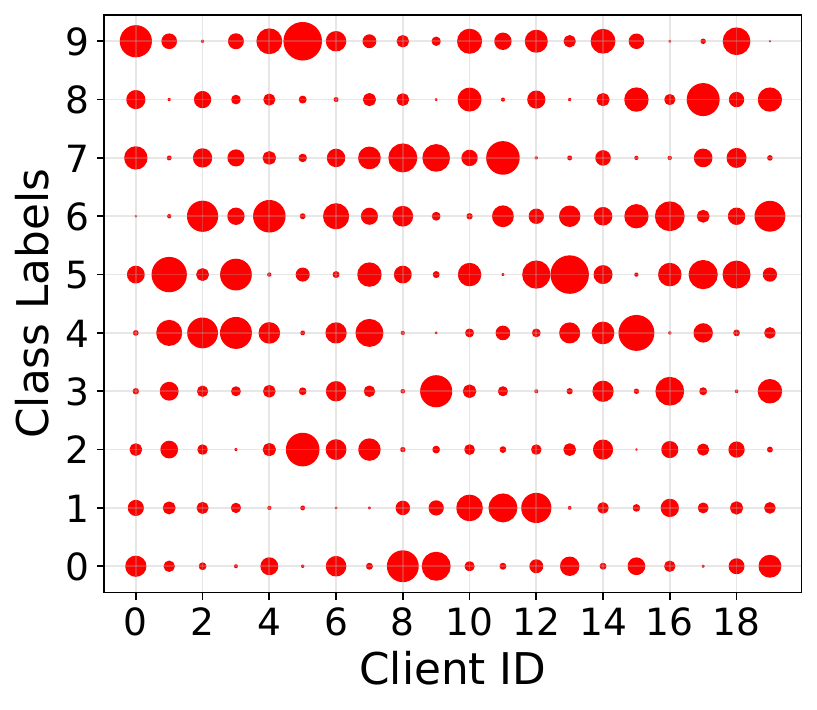}}
	\subfigure[$\alpha=1.0$]{
		\includegraphics[width=0.45\linewidth]{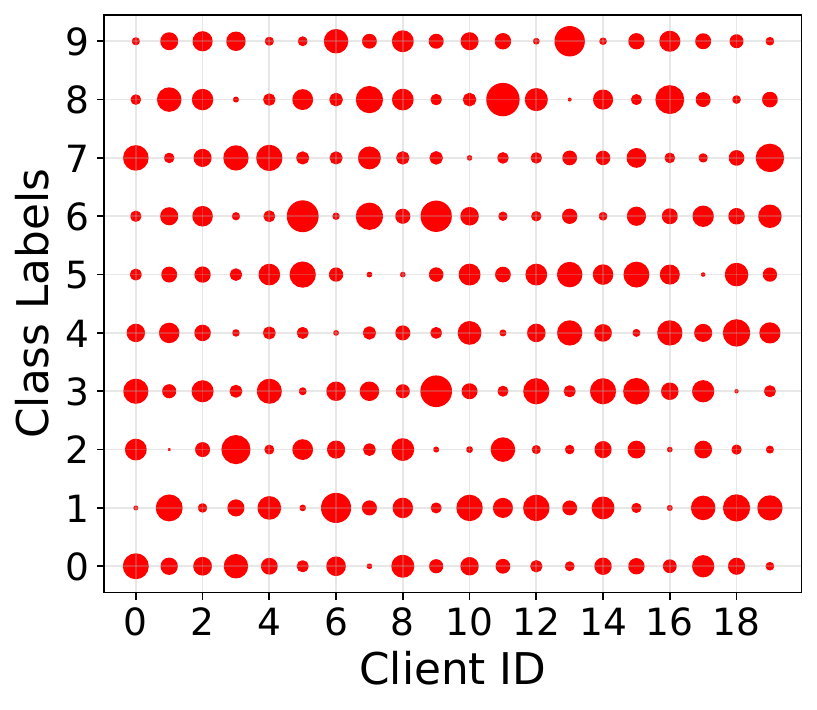}}
	
	\caption{Visualization of data partitioning in Dirichlet non-IID scenarios with different $\alpha$.}
	\label{fig:dirichlet example}
\end{figure}
To facilitate intuitive understanding, we utilize 20 clients on the 10-classification dataset to visualize the data distribution of clients with different $\alpha$ values.  As shown in Figure \ref{fig:dirichlet example}, the horizontal axis represents the client ID, and the vertical axis represents the data class label index.  Red dots represent the data assigned to clients.  The larger the dot is, the more data the client has in this class.  When $\alpha$ is small (e.g., $\alpha=0.01$), the overall data distributions of clients vary greatly.  However, the variety of client data distribution is low, and it is easy to have clients with very similar data distributions.  As the $\alpha$ increases, the difference in data distribution among clients gradually decreases while the variety of client data distribution increases.

\section{Additional Comparison with SOTAs}
Parallel with our work, we find two PFL works pFedGate \cite{chen2023efficient} and perFedMask \cite{setayesh2022perfedmask} that also propose to use mask matrices. To further enhance the persuasiveness of our work, we add experiments to compare with them.

Since our primary focus is on the accuracy, we choose hyperparameter settings that yield the best accuracy for each method. For instance, we set $s=1$ in pFedGate and $\nu=1$ in PerFedMask. Table~\ref{pathological noniid} presents the test accuracy achieved by the different methods. Notably, FedCAC outperforms the other methods in terms of accuracy. This improvement can be attributed to FedCAC's ability to facilitate more refined collaboration by considering data distribution similarity and parameter sensitivity, thereby enhancing robustness in non-IID scenarios.

Furthermore, we would like to clarify that although all aforementioned methods utilize masks, their purposes and functionalities differ significantly. While pFedGate utilizes masks for personalized model adaptation and PerFedMask employs masks to freeze layers for computation reduction, our method utilizes masks to control parameter collaboration mode and represent client data distribution. 

\renewcommand\arraystretch{0.8}
\begin{table}[tb]
	\vskip -0.0in
	\begin{center}
				\begin{tabular}{lcccr}
					\toprule
					Methods & CIFAR-10 & CIFAR-100 \\
					\midrule
					pFedGate (ICML 2023) & 89.15 $\pm$ 0.76 & 92.30 $\pm$ 0.72 \\
					PerFedMask (ICLR 2023) & 89.20 $\pm$ 0.65 & 92.12 $\pm$ 0.56 \\
					\midrule
					FedCAC & \textbf{89.77 $\pm$ 1.14} & \textbf{93.05 $\pm$ 0.90}  \\
					\bottomrule
				\end{tabular}
	\end{center}

 \caption{Comparison results under Pathological non-IID.}
 \label{pathological noniid}
	\vskip -0.0in
\end{table}

\end{document}